\documentclass[10pt,twocolumn,letterpaper]{article}

\usepackage{cvpr}              

\usepackage{graphicx}
\usepackage{amsmath}
\usepackage{amssymb}
\usepackage{booktabs}
\usepackage{multirow}

\usepackage[pagebackref,breaklinks,colorlinks]{hyperref}

\usepackage[capitalize]{cleveref}
\crefname{section}{Sec.}{Secs.}
\Crefname{section}{Section}{Sections}
\Crefname{table}{Table}{Tables}
\crefname{table}{Tab.}{Tabs.}

\def\cvprPaperID{13} 
\def\confName{CVPR}
\def\confYear{2022}

\begin{document}

\title{Transformer for Single Image Super-Resolution}

\author{Zhisheng Lu$^{1\dagger}$,\,Juncheng Li$^{2\dagger}$,\,Hong Liu$^1\thanks{Corresponding author $\dagger$Co-first authors}$,\,Chaoyan Huang$^3$,\,Linlin Zhang$^1$,\,Tieyong Zeng$^2$\\
$^{1}$Peking University Shenzhen Graduate School
$^{2}$The Chinese University of Hong Kong\\
$^{3}$Nanjing University of Posts and Telecommunications\\
{\tt\small \{zhisheng\_lu, hongliu, catherinezll\}@pku.edu.cn}\\
{\tt\small cvjunchengli@gmail.com, Huangchy2020@163.com, zeng@math.cuhk.edu.hk}
}

\maketitle

\begin{abstract}
Single image super-resolution (SISR) has witnessed great strides with the development of deep learning. However, most existing studies focus on building more complex networks with a massive number of layers. Recently, more and more researchers start to explore the application of Transformer in computer vision tasks. However, the heavy computational cost and high GPU memory occupation of the vision Transformer cannot be ignored. In this paper, we propose a novel Efficient Super-Resolution Transformer (ESRT) for SISR. ESRT is a hybrid model, which consists of a Lightweight CNN Backbone (LCB) and a Lightweight Transformer Backbone (LTB). Among them, LCB can dynamically adjust the size of the feature map to extract deep features with a low computational costs. LTB is composed of a series of Efficient Transformers (ET), which occupies a small GPU memory occupation, thanks to the specially designed Efficient Multi-Head Attention (EMHA). Extensive experiments show that ESRT achieves competitive results with low computational cost. Compared with the original Transformer which occupies 16,057M GPU memory, ESRT only occupies 4,191M GPU memory. All codes are available at \url{https://github.com/luissen/ESRT}.
\end{abstract}

\section{Introduction}
{S}{ingle} image super-resolution (SISR) aims at recovering a super-resolution (SR) image from its degraded low-resolution (LR) counterpart, which is a useful technology to overcome resolution limitations in many applications. However, it still is an ill-posed problem since there exist infinite HR images. To address this issue, numerous deep neural networks have been proposed~\cite{srcnn,vdsr,edsr,msrn,he2018cascaded,wang2018resolution,zhang2018residual,Wang2019ResolutionAwareNF,li2021beginner}. Although these methods have achieved outstanding performance, they cannot be easily utilized in real applications due to high computation cost and memory storage. To solve this problem, many recurrent networks and lightweight networks have been proposed, such as DRCN~\cite{drcn}, SRRFN~\cite{li2019lightweight}, IMDN~\cite{imdn}, IDN~\cite{idn}, CARN~\cite{CARN},ASSLN~\cite{zhang2021aligned}, MAFFSRN~\cite{muqeet2020multi}, and RFDN~\cite{liu2020residual}. All these models concentrate on constructing a more efficient network structure, but the reduced network capacity will lead to poor performance. 

\begin{figure}[t]
\centering
\includegraphics[scale=0.14]{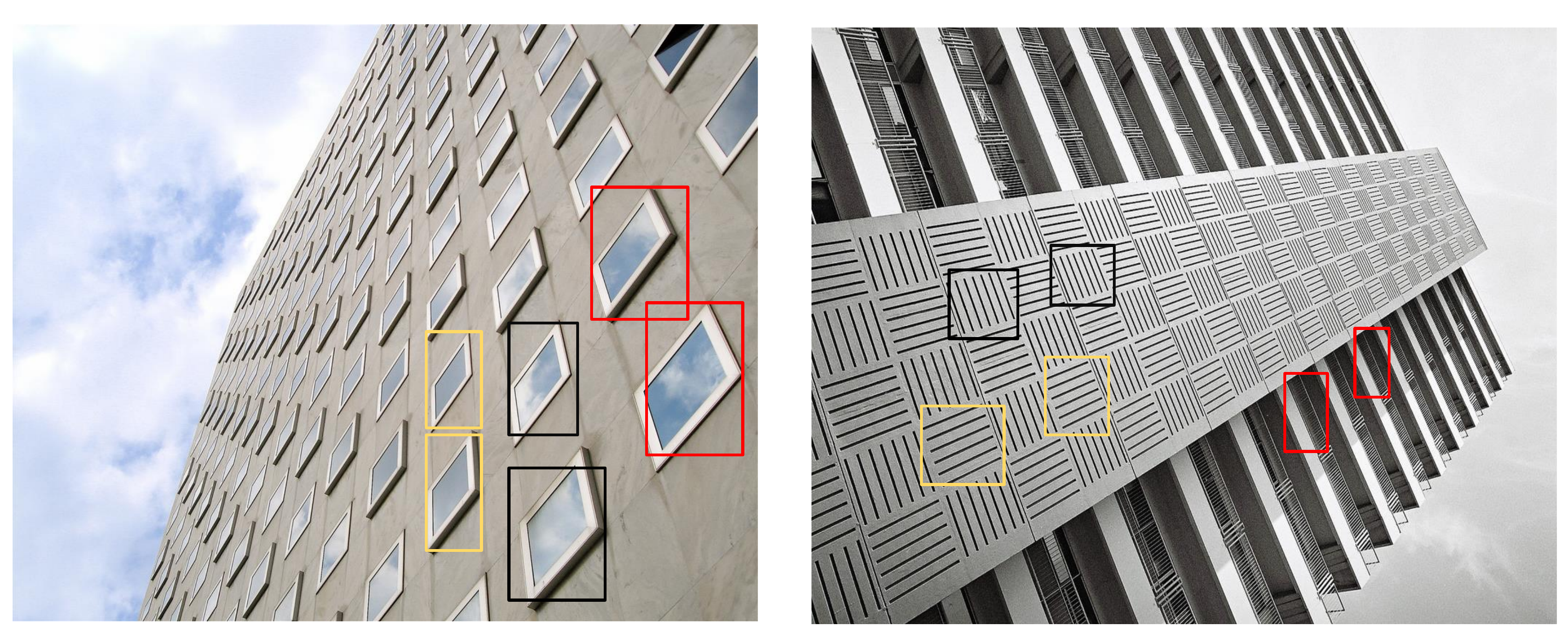}
\caption{Examples of similar patches in images. These similar patches can help restore details from each other.}
\label{fig:reason}
\vspace{-0.3cm}
\end{figure}

As Figure~\ref{fig:reason} shows, the inner areas of the boxes with the same color are similar to each other. Therefore, these similar image patches can be used as reference images for each other, so that the texture details of the certain patch can be restored with reference patches. Inspired by this, we introduce the Transformer into the SISR task since it has a strong feature expression ability to model such a long-term dependency in the image. In other words, we aim to explore the feasibility of using Transformer in the lightweight SISR task. In recent years, some Vision-Transformer~\cite{liu2021swin,ViT} have been proposed for computer vision tasks. However, these methods often occupy heavy GPU memory, which greatly limits their flexibility and application scenarios. Moreover, these methods cannot be directly transferred to SISR since the image restoration task often take a larger resolution image as input, which will take up huge memory.

To solve the aforementioned problems, an Efficient Super-Resolution Transformer (ESRT) is proposed to enhance the ability of SISR networks to capture the long-distance context-dependence while significantly decreasing the GPU memory cost. It is worth noting that ESRT is a hybrid architecture, which uses a ``CNN+Transformer" pattern to handle the small SR dataset. Specifically, ESRT can be divided into two parts: Lightweight CNN Backbone (LCB) and Lightweight Transformer Backbone (LTB). For LCB, we consider more on reducing the shape of the feature map in the middle layers and maintaining a deep network depth to ensure large network capacity. Inspired by the high-pass filter, we design a High-frequency Filtering Module (HFM) to capture the texture details of the image. With the aid of HFM, a High Preserving Block (HPB) is proposed to extract the potential features efficiently by size variation. For feature extraction, a powerful Adaptive Residual Feature Block (ARFB) is proposed as the basic feature extraction unit with the ability to adaptively adjust the weight of the residual path and identity path. In LTB, an efficient Transformer (ET) is proposed, which use the specially designed Efficient Multi-Head Attention (EMHA) mechanism to decrease the GPU memory consumption. It is worth noting that EMHA just considers the relationship between image blocks in a local region since the pixel in SR image is commonly related to its neighbor pixels. Even though it is a local region, it is much wider than a regular convolution and can extract more useful context information. Therefore, ESRT can learn the relationship between similar local blocks efficiently, making the super-resolved region have more references. The main contributions are as follows
\begin{itemize}  
    \item We propose a Lightweight CNN Backbone (LCB), which use High Preserving Blocks (HPBs) to dynamically adjust the size of the feature map to extract deep features with a low computational cost.
    \item We propose a Lightweight Transformer Backbone (LTB) to capture long-term dependencies between similar patches in an image with the help of the specially designed Efficient Transformer (ET) and Efficient Multi-Head Attention (EMHA) mechanism.
    \item A novel model called Efficient SR Transformer (ESRT) is proposed to effectively enhance the feature expression ability and the long-term dependence of similar patches in an image, so as to achieve better performance with low computational cost.
\end{itemize}

\section{Related Works}
\subsection{CNN-based SISR Models}
Recently, many CNN-base models have been proposed for SISR. For example, SRCNN~\cite{srcnn} first introduces the deep CNN into SISR and achieves promising results. EDSR~\cite{edsr} optimizes the residual block by removing unnecessary operations and expanding the model size. RCAN~\cite{RCAN} proposes a deep residual network with residual-in-residual architecture and channel attention mechanism. SAN~\cite{san} presents a second-order attention network to enhance the feature expression and feature correlation learning. IDN~\cite{idn} compresses the model size by using the group convolution and combining short-term and long-term features. IMDN~\cite{imdn} improves the architecture of IDN and introduces the information multi-distillation blocks to extract the hierarchical features effectively. LatticeNet~\cite{latticenet} designs the lattice block that simulates the realization of Fast Fourier Transformation with the butterfly structure. Although these models achieved competitive results, they are pure CNN-based model. This means that they can only extract local features and cannot learn the global information, which is not conducive to the restoration of texture details.

\begin{figure*}[ht]
\centering
\includegraphics[scale=0.46]{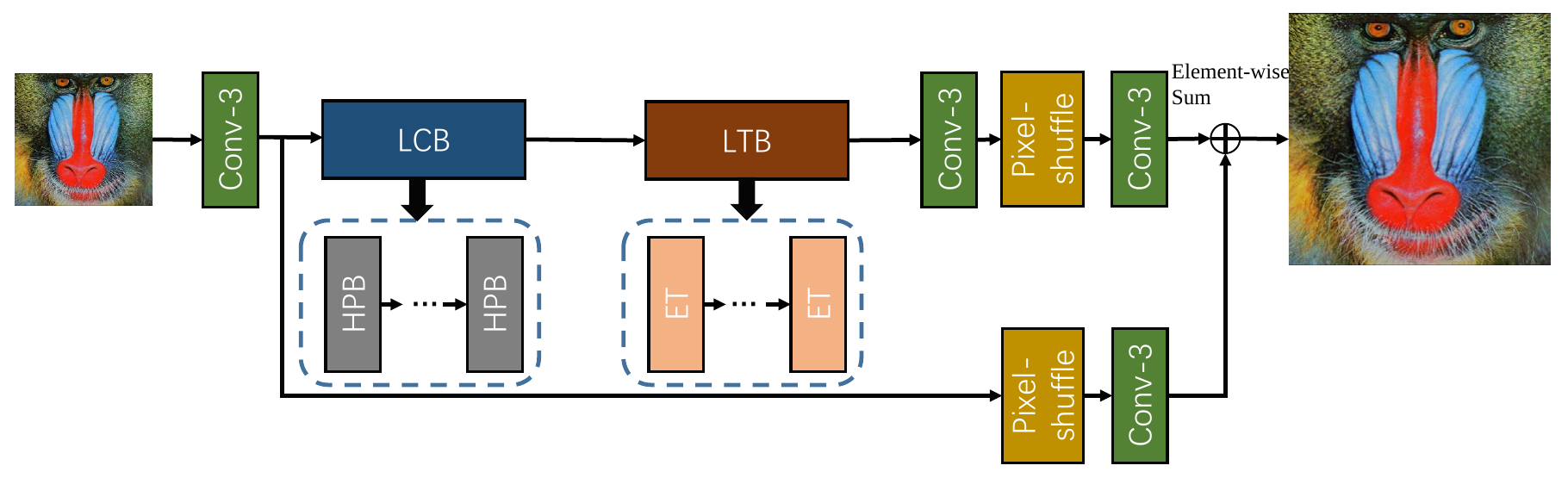}
\caption{The architecture of the proposed Efficient Super-Resolution Transformer. Among them, LCB, LTB, HPB, and ET stand for the Lightweight CNN Backbone, the Lightweight Transformer Backbone, high preserving block, and efficient Transformers, respectively.}
\label{fig:archi}
\vspace{-0.3cm}
\end{figure*}

\subsection{Vision Transformer}
The breakthroughs of Transformer in NLP has leaded to a great interest in the computer vision community. The key idea of Transformer is ``self-attention", which can capture long-term information between sequence elements. By adapting Transformer in vision tasks, it has been successfully applied in image recognition~\cite{ViT,deit,li2021localvit}, object detection~\cite{detr,deformabledetr}, and low-level image processing~\cite{IPT,ttsr}. Among them, ViT~\cite{ViT} is the first work to replace the standard convolution with Transformer. To produce the sequence elements, ViT flattened the 2D image patches in a vector and fed them into the Transformer. IPT~\cite{IPT} used a novel Transformer-based network as the pre-trained model for low-level image restoration tasks. SwinIR~\cite{liang2021swinir} introduced Swin Transformer~\cite{liu2021swin} into SISR and show the great promise of Transformer in SISR. Although these methods achieved promising results, they require a lot of training data and need heavy GPU memory to train the model, which is not suitable for practical applications. Hence, we aim to explore a more efficient vision-Transformer for SISR.

\section{Efficient Super-Resolution Transformer}
As shown in Figure~\ref{fig:archi}, Efficient Super-Resolution Transformer (ESRT) mainly consists of four parts: shallow feature extraction, Lightweight CNN Backbone (LCB), Lightweight Transformer Backbone (LTB), and image reconstruction. Define $I_{LR}$ and $I_{SR}$ as the input and output of ESRT, respectively. Firstly, we extract the shallow feature from $I_{LR}$ with a convolutional layer
\begin{equation}\label{equ:eq1}
F_0 = f_s(I_{LR}),
\end{equation}
where $f_s$ denotes the shallow feature extraction layer. $F_0$ is the extracted shallow feature, which is then used as the input of LCB with several High Preserving Blocks (HPBs)
\begin{equation}\label{equ:eq2}
F_n = \zeta^{n}(\zeta^{n-1}(...(\zeta^{1}(F_0)))),
\end{equation}
where $\zeta^{n}$ denotes the mapping of $n$-th HPB and $F_n$ represents the output of $n$-th HPB. All outputs of HPB are concatenated to be sent to LTB with several Efficient Transformers (ETs) to fuse these intermediate features
\begin{equation}
    F_d = \phi^{n}(\phi^{n-1}(...(\phi^{1}([F_1, F_2,...,F_n])))),
    \label{equ:eq3}
\end{equation}
where $F_d$ is the output of LTB and $\phi$ stands for the operation of ET. Finally, $F_d$ and $F_0$ are simultaneously fed into the reconstruction module to get the SR image $I_{SR}$
\begin{equation}
    I_{SR} = f(f_p(f(F_d))) + f(f_p(F_0)),
    \label{equ:eq4}
\end{equation}
where $f$ and $f_p$ stand for the convolutional layer and Pixel-Shuffle layer, respectively.

\subsection{Lightweight CNN Backbone (LCB)}
\label{sec:lnmb}
The role of Lightweight CNN Backbone (LCB) is to extract potential SR features in advance, so that the model has the initial ability of super-resolution. According to Figure~\ref{fig:archi}, we can observe that LCB is mainly composed of a series of High Preserving Blocks (HPBs). 

\textbf{High Preserving Block (HPB).} Previous SR networks usually keep the spatial resolution of feature maps unchanged during processing. In this work, in order to reduce the computational cost, a novel High Preserving Block (HPB) is proposed to reduce the resolution of processing features. However, the reduction of the size of feature maps always leads to the loss of image details, which causes visually unnatural SR images. To solve this problem, in HPB, we creatively preserve the High-frequency Filtering Module (HFM) and Adaptive Residual Feature Block (ARFB).

\begin{figure}[t]
\includegraphics[scale=0.38]{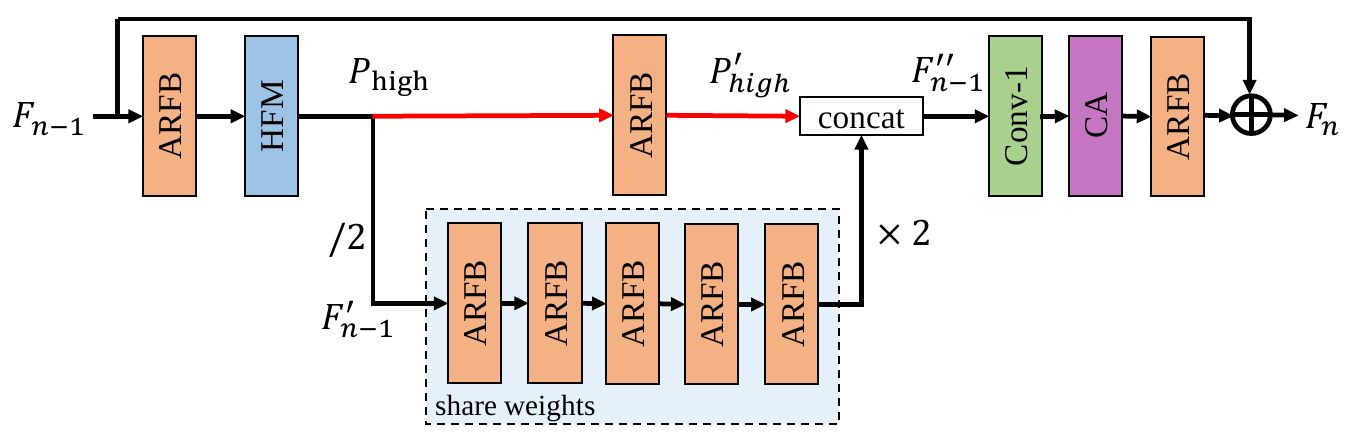}
\caption{The architecture of the proposed High Preserving Block (HPB), which mainly consists of High-frequency Filtering Module (HFM) and Adaptive Residual Feature Blocks (ARFBs).}
\label{fig:hlsb}
\vspace{-0.3cm}
\end{figure}

As shown in Figure~\ref{fig:hlsb}, an ARFB is first use to extract $F_{n-1}$ as the input features for HFM. Then, HFM is used to calculate the high-frequency information (marked as $P_{high}$) of the features. After the $P_{high}$ is obtained, we reduce the size of the feature map to reduce computational cost and feature redundancy. The downsampled feature maps are denoted as $F^{'}_{n-1}$. For $F^{'}_{n-1}$, several ARFBs are utilized to explore the potential information for completing the SR image. It is worth noting that these ARFBs share weights to reduce parameters. Meanwhile, a single ARFB is used to process the $P_{high}$ to align the feature space with $F_{n-1}^{'}$. After feature extraction, $F_{n-1}^{'}$ is upsampled to the original size by bilinear interpolation. After that, we fuse the $F_{n-1}^{'}$ with $P_{high}^{'}$ for preserving the initial details and obtain the feature $F_{n-1}^{''}$. This operation can be expressed as
\begin{equation}\label{equ:eq5}
F_{n-1}^{''} = [f_{a}(P_{high}), \uparrow f_{a}^{\circlearrowright^5}(\downarrow F_{n-1}^{'})],
\end{equation}
where $\uparrow$ and $\downarrow$ denote the upsampling and downsampling operations, respectively. $f_{a}$ denotes the operation of ARFB. To achieve good trade-off between the model size and performance, we use five ARFBs in this part according to ablation studies and define it as $f_{a}^{\circlearrowright^5}$.

For $F_{n-1}^{''}$, as it is concatenated by two features, a $1\times 1$ convolution layer is used to reduce the channel number. Then, a channel attention module~\cite{senet} is employed to highlight channels with high activated values. Finally, an ARFB is used to extract the final features and the global residual connection is proposed to add the original features $F_{n-1}$ to $F_{n}$. The goal of this operation is to learn the residual information from the input and stabilize the training.

\begin{figure}
\begin{center}
\includegraphics[scale=0.4]{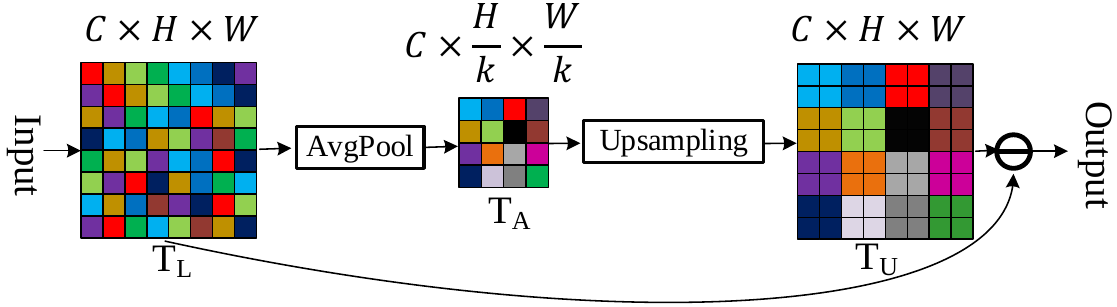}
\end{center}
\caption{The schematic diagram of the proposed HFM module.}
\label{fig:hfm}
\vspace{-0.3cm}
\end{figure}

\begin{figure}
\begin{center}
\includegraphics[scale=0.3]{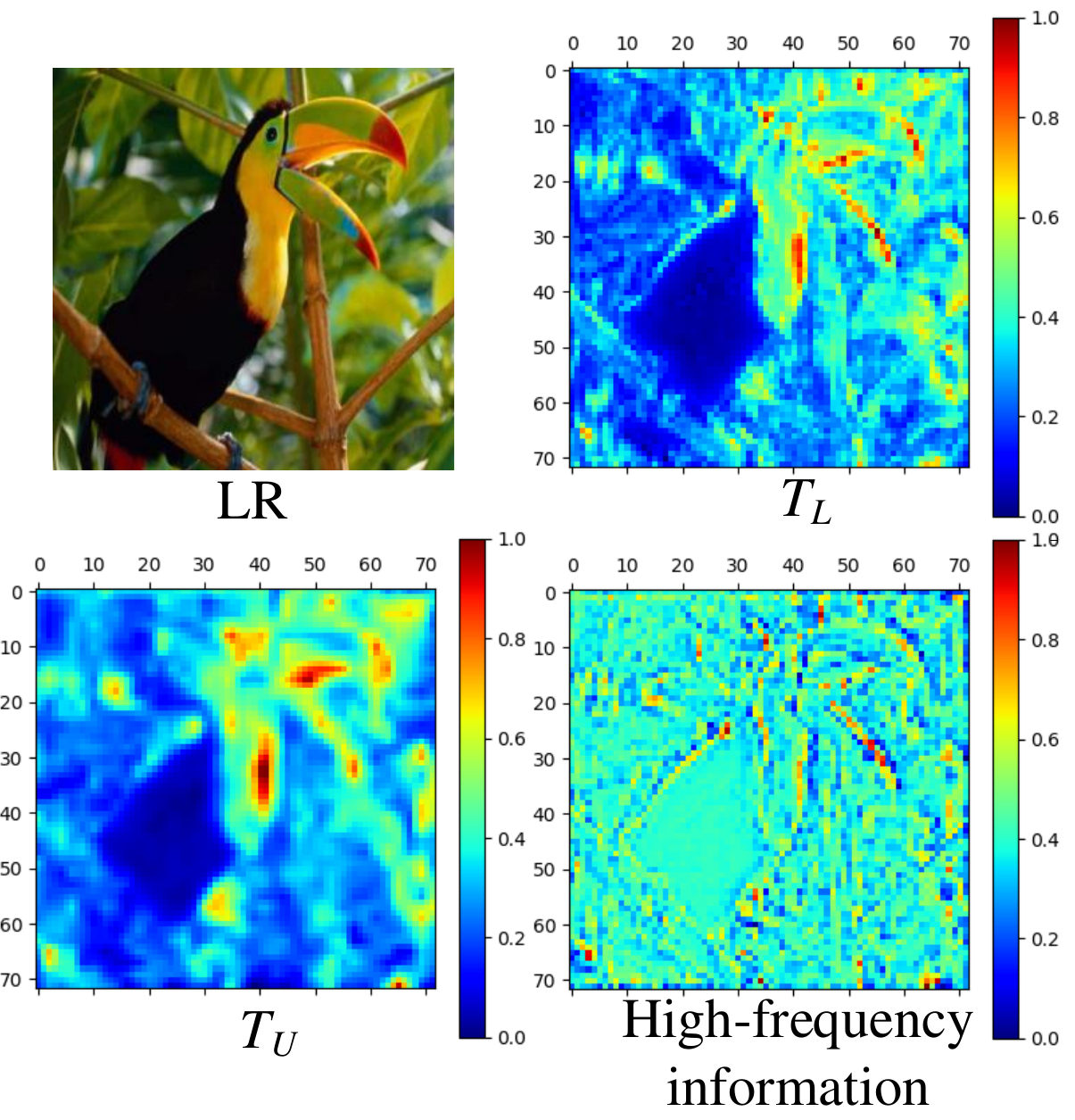}
\end{center}
\caption{Visual activation maps of $T_L$, $T_U$, and obtained high-frequency information. Best viewed in color.}
\label{fig:vis}
\vspace{-0.3cm}
\end{figure}

\subsection{\textbf{High-frequency Filtering Module (HFM)}}
Since the Fourier Transform is difficult to embed in CNN, a differentiable HFM is proposed in this work. The target of HFM is to estimate the high-frequency information of the image from the LR space. As shown in Figure~\ref{fig:hfm}, assuming the size of the input feature map $T_L$ is $C\times H \times W$, an average pooling layer is first applied to $T_L$:
\begin{equation}\label{equ:eq5}
T_{A} = avgpool(T_L, k),
\end{equation}
where $k$ denotes the kernel size of the pooling layer and the size of the intermediate feature map $T_A$ is $C\times \frac{H}{k} \times \frac{W}{k}$. Each value in $T_A$ can be viewed as the average intensity of each specified small area of $T_L$. After that, $T_A$ is upsampled to get a new tensor $T_U$ of size $C\times H\times W$. $T_U$ is regarded as an expression of the average smoothness information compared with the original $T_L$. Finally, $T_U$ is element-wise subtracted from $T_L$ to obtain the high-frequency information.

The visual activation maps of $T_L$, $T_U$, and high-frequency information are also shown in Figure~\ref{fig:vis}. It can be observed that the $T_U$ is more smooth than $T_L$ as it is the average information of the $T_L$. Meanwhile, the high-frequency information retains the details and edges of the feature map before downsampling. Therefore, it is essential to save these information.

\begin{figure}
      \centering
      \includegraphics[scale=0.45]{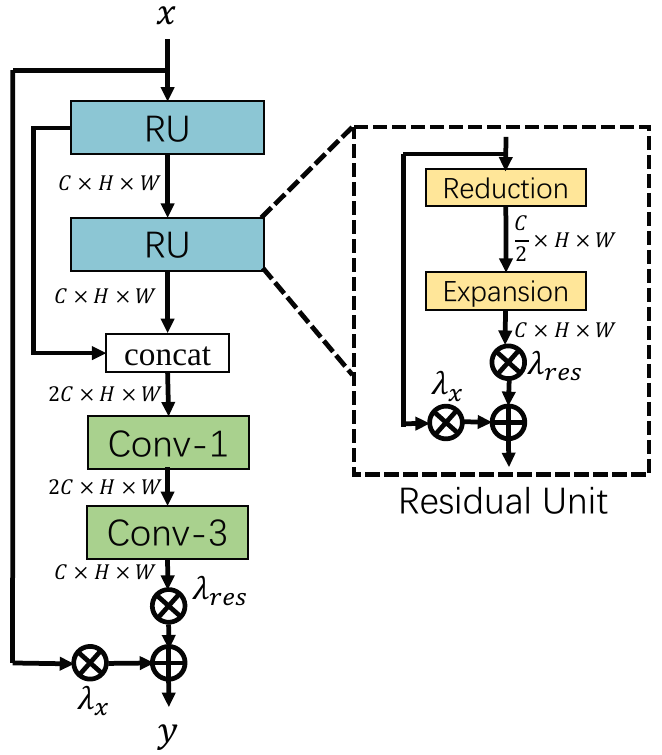}
      \caption{The complete architecture of the proposed ARFB.}
      \label{fig:arfb}
      \vspace{-0.3cm}
\end{figure}

\begin{figure*}[t]
      \centering
      \includegraphics[scale=0.45]{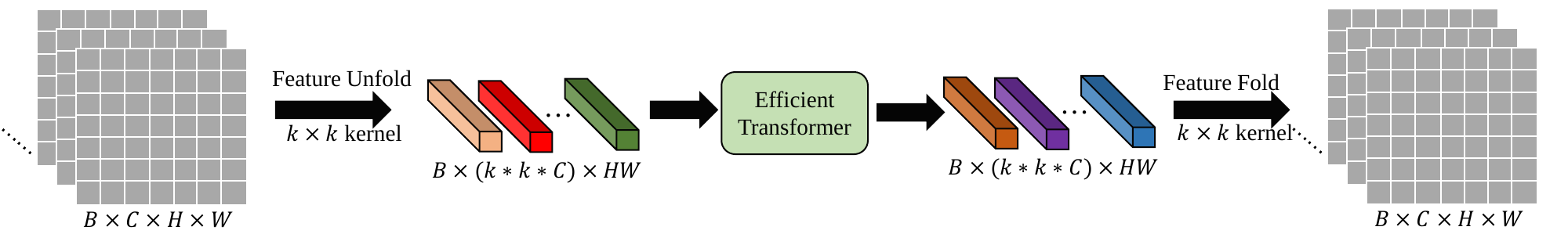}
      \caption{The complete pre- and post-processing for the Efficient Transformer (ET). Specifically, we use the unfolding technique to split the feature maps into patches and use the fold operation to reconstruct the feature map.}
      \label{fig:preprocess}
      \vspace{-0.3cm}
\end{figure*}

\subsubsection{\textbf{Adaptive Residual Feature Block (ARFB)}}
As explored in ResNet~\cite{he2016deep} and VDSR~\cite{vdsr}, when the depth of the model grows, the residual architecture can mitigate the gradient vanishing problem and augment the representation capacity of the model. Inspired by them, a Adaptive Residual Feature Block (ARFB) is proposed as the basic feature extraction block. As shown in Figure~\ref{fig:arfb}, ARFB contains two Residual Units (RUs) and two convolutional layers. To save memory and the number of parameters, RU is made up of two modules: Reduction and Expansion. For Reduction, the channels of the feature map are reduced by half and recovered in Expansion. Meanwhile, a residual scaling with adaptive weights (RSA) is designed to dynamically adjust the importance of residual path and identity path. Compared with fixed residual scaling, RSA can improve the flow of gradients and automatically adjust the content of the residual feature maps for the input feature map. Assume that $x_{ru}$ is the input of RU, the process of RU can be formulated as:
\begin{equation}
    y_{ru} = \lambda_{res}\cdot f_{ex}(f_{re}(x_{ru})) + \lambda_{x}\cdot x,
    \label{equ:eq6}
\end{equation}
where $y_{ru}$ is the output of RU, $f_{re}$ and $f_{ex}$ represent the Reduction and Expansion operations, $\lambda_{res}$ and $\lambda_{x}$ are two adaptive weights for two paths, respectively. These two operations use $1\times 1$ convolutional layers to change the number of channels to achieve the functions of reduction and expansion. Meanwhile, the outputs of two RUs are concatenated followed by a $1\times 1$ convolutional layer to fully utilize the hierarchical features. In the end, a $3\times 3$ convolutional layer is adopted to reduce the channels of the feature map and extract valid information from the fused features.

\subsection{Lightweight Transformer Backbone (LTB)}
In SISR, similar image blocks within the image can be used as reference images to each other, so that the texture details of the current image block can be restored with reference to other image blocks, which is proper to use Transformer. However, previous variants of vision Transformer commonly need heavy GPU memory cost, which hinders the development of Transformer in the vision area. In this paper, we propose a Lightweight Transformer Backbone (LTB). LTB is composed of specially designed Efficient Transformers (ETs), which can capture the long-term dependence of similar local regions in the image at a low computational cost. 

\subsubsection{\textbf{Pre- and Post-processing for ET}}
The standard Transformer takes a 1-D sequence as input, learning the long-distance dependency of the sequence. However, for the vision task, the input is always a 2-D image. The common way to turn a 2-D image into a 1-D sequence is to sort the pixels in the image one by one. However, this method will lose the unique local correlation of the image, leading to sub-optimal performance. In ViT~\cite{ViT}, the 1-D sequence is generated by non-overlapping block partitioning, which means there is no pixel overlap between each block. According to our experiments, these pre-processing methods are not suitable for SISR. Therefore, a novel processing way is proposed to handle the feature maps.

As shown in Figure~\ref{fig:preprocess}, we use the unfolding technique to split the feature maps into patches and each patch is considered as a ``word". Specifically, the feature maps $F_{ori} \in \mathbb{R}^{C\times H\times W}$ are unfolded (by $k\times k$ kernel) into a sequence of patches, i.e., $F_{p_i} \in \mathbb{R}^{k^2\times C}$, $i=\{1,...,N\}$, where $N=H\times W$ is the amount of patches. Here, the learnable position embeddings are eliminated for each patch since the ``Unfold" operation automatically reflects the position information for each patch. After that, those patches $F_p$ are directly sent to the ET. The output of ET has the same shape as the input and we use the ``Fold" operation to reconstruct feature maps.

\subsubsection{\textbf{Efficient Transformer (ET)}}
\begin{figure}[t]
  \centering
  \includegraphics[scale=0.3]{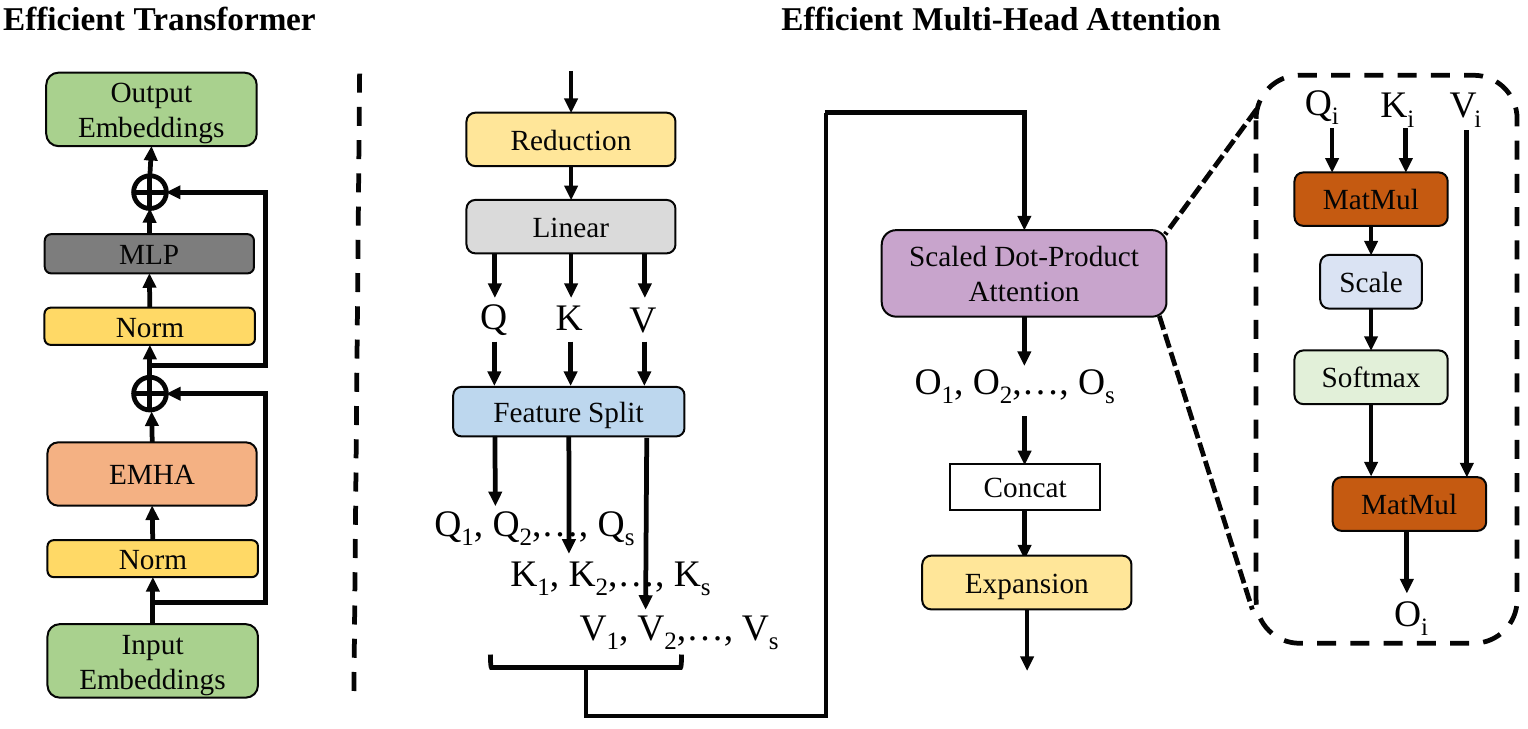}
  \caption{Architecture of Efficient Transformer. EMHA is the Efficient Multi-Head Attention. MatMul is the matrix multiplication.}
  \label{fig:erst}
  \vspace{-0.3cm}
\end{figure}

For simplicity and efficiency, like ViT~\cite{ViT}, ET only uses the encoder structure of the standard Transformer. As shown in Figure~\ref{fig:erst}, in the encoder of ET, there consists of an Efficient Multi-Head Attention (EMHA) and an MLP. Meanwhile, layer-normalization~\cite{ln} is employed before every block, and the residual connection is also applied after each block. Assume the input embeddings are $E_i$, the output embeddings $E_o$ can be obtained by
\begin{equation}\label{equ:eq6}
\begin{split}
E_{m1} & = EMHA(Norm(E_i))+E_i,\\
E_o & = MLP(Norm(E_{m1})) +E_{m1},
\end{split}
\end{equation}
where $E_o$ is the output of the ET, $EMHA(\cdot)$ and $MLP(\cdot)$ represent the EMHA and MLP operations, respectively.

\begin{table*}[t]
  \centering
  \small
  \setlength{\tabcolsep}{2.5mm}
  \renewcommand{\arraystretch}{0.88}
	\begin{tabular}{l|c|c|c|c|c|c|c}
		\toprule[1pt]
		\multirow{2}{*}{Method} & \multirow{2}{*}{Scale} & \multirow{2}{*}{Params} & Set5 & Set14 & BSD100 & Urban100 & Manga109 \\
		\cline{4-8}
		& & & PSNR / SSIM & PSNR / SSIM & PSNR / SSIM & PSNR / SSIM & PSNR / SSIM \\
    \hline
    		
		VDSR~\cite{vdsr} & \multirow{9}{*}{$\times 3$} & 666K & 33.66 / 0.9213 & 29.77 / 0.8314 & 28.82 / 0.7976 & 27.14 / 0.8279 & 32.01 / 0.9340 \\
		
		MemNet~\cite{memnet} &  & 678K & 34.09 / 0.9248 & 30.00 / 0.8350 & 28.96 / 0.8001 & 27.56 / 0.8376 & 32.51 / 0.9369\\
		
		EDSR-baseline~\cite{edsr} &  & 1,555K & 34.37 / 0.9270 & 30.28 / 0.8417 & 29.09 / 0.8052 & 28.15 / 0.8527 & 33.45 / 0.9439 \\
		
		SRMDNF~\cite{srmdnf} &  & 1,528K & 34.12 / 0.9254 & 30.04 / 0.8382 & 28.97 / 0.8025 & 27.57 / 0.8398 & 33.00 / 0.9403 \\
		
		CARN~\cite{CARN} &  & 1,592K & 34.29 / 0.9255 & 30.29 / 0.8407 & 29.06 / 0.8034 & 28.06 / 0.8493 & 33.50 / 0.9440 \\
		
    IMDN~\cite{imdn} &  & 703K & 34.36 / 0.9270 & 30.32 / 0.8417 & 29.09 / 0.8046 & 28.17 / 0.8519 & 33.61 / 0.9445 \\

    RFDN-L~\cite{liu2020residual} &  & 633K & \underline{34.47} / \underline{0.9280} &  30.35 / 0.8421 & 29.11 / 0.8053 & 28.32 / 0.8547  & \underline{33.78} / \underline{0.9458} \\

    MAFFSRN~\cite{muqeet2020multi} & & 807K & 34.45 / 0.9277 & \underline{30.40} / \underline{0.8432} & \underline{29.13} / \underline{0.8061} & 28.26 / \underline{0.8552} &  - / -\\
    
    LatticeNet~\cite{latticenet} &  & 765K & \textbf{34.53} / \textbf{0.9281} & 30.39 / 0.8424 & \textbf{29.15} / 0.8059 & \underline{28.33} / 0.8538 & - / -\\
		
		\textbf{ESRT(ours)} &  & 770K & 34.42 / 0.9268 & \textbf{30.43} / \textbf{0.8433} & \textbf{29.15} / \textbf{0.8063} & \textbf{28.46} / \textbf{0.8574} & \textbf{33.95} / \textbf{0.9455} \\
		
		\hline
		
		VDSR~\cite{vdsr} & \multirow{9}{*}{$\times 4$} & 666K & 31.35 / 0.8838 & 28.01 / 0.7674 & 27.29 / 0.7251 & 25.18 / 0.7524 & 28.83 / 0.8870 \\
		
		MemNet~\cite{memnet} &  & 678K & 31.74 / 0.8893 & 28.26 / 0.7723 & 27.40 / 0.7281 & 25.50 / 0.7630 & 29.42 / 0.8942 \\
		
		EDSR-baseline~\cite{edsr} &  & 1,518K & 32.09 / 0.8938 &  28.58 / 0.7813 & 27.57 / 0.7357 & 26.04 / 0.7849 & 30.35 / 0.9067 \\
		
		SRMDNF~\cite{srmdnf} &  & 1,552K & 31.96 / 0.8925 & 28.35 / 0.7787 & 27.49 / 0.7337 & 25.68 / 0.7731 & 30.09 / 0.9024\\
		
    CARN~\cite{CARN} &  & 1,592K & 32.13 / 0.8937 & 28.60 / 0.7806 &  27.58 / 0.7349 & 26.07 / 0.7837  & 30.47 / 0.9084 \\
		
    IMDN~\cite{imdn} &  & 715K & 32.21 / 0.8948 & 28.58 / 0.7811 & 27.56 / 0.7353 & 26.04 / 0.7838 & 30.45 / 0.9075 \\

    RFDN-L~\cite{liu2020residual} &  & 643K & \underline{32.28} / \underline{0.8957} & 28.61 / 0.7818 & 27.58 / 0.7363 & 26.20 / 0.7883 & \underline{30.61} / \underline{0.9096} \\

    MAFFSRN~\cite{muqeet2020multi} & & 830K & 32.20 / 0.8953 & 26.62 / 0.7822 & 27.59 / 0.7370 & 26.16 / \underline{0.7887} & - / - \\
    
    LatticeNet~\cite{latticenet} &  & 777K & \textbf{32.30} / \textbf{0.8962} & \underline{28.68} / \underline{0.7830} & \underline{27.62} / \underline{0.7367} & \underline{26.25} / 0.7873 & - / -\\
		
		\textbf{ESRT (ours)} &  & 751K & 32.19 / 0.8947 & \textbf{28.69} / \textbf{0.7833} & \textbf{27.69} / \textbf{0.7379} & \textbf{26.39} / \textbf{0.7962} & \textbf{30.75} / \textbf{0.9100} \\
		\bottomrule[1pt]
		
  \end{tabular}
  \caption{Quantitative comparison with SISR models. The Best and the second-best results are \textbf{highlighted} and \underline{underlined}, respectively.}
  \label{tab:psnr-ssim}
  \vspace{-0.3cm}
\end{table*}

\textbf{Efficient Multi-Head Attention (EMHA)}. As shown in Figure~\ref{fig:erst}, there are several modifications in EMHA to make EMHA more efficient and occupy lower GPU memory cost compared with the original MHA~\cite{TR}. Assume the shape of the input embedding $E_i$ is $B\times C \times N$. Firstly, a Reduction layer is used to reduce the number of channels by half ($B\times C_1 \times N, C_1 = \frac{C}{2}$). After that, a linear layer is adopted to project the feature map into three elements: $Q$ (query), $K$ (keys), and $V$ (values). As employed in Transformer, we linearly project the $Q$, $K$, $V$, and $m$ times to perform the multi-head attention. $m$ is the number of heads. Next, the shape of three elements is reshaped and permuted to $B\times m \times N \times \frac{C_1}{m}$. In original MHA, $Q$, $K$, $V$ are directly used to calculate the self-attention with large-scale matrix multiplication, which cost huge memory. Assume $Q$ and $K$ calculate the self-attention matrix with shape $B\times m \times N \times N$. Then this matrix computes the self-attention with $V$, the dimension in $3$-th and $4$-th are $N\times N$. For SISR, the images usually have high resolution, causing that $N$ is very large and the calculation of self-attention matrix consumes a lot of GPU memory cost and computational cost. To address this issue, a Feature Split (FS) Module is used to split $Q$, $K$, and $V$ into $s$ equal segments with splitting factor $s$ since the predicted pixels in super-resolved images often only depend on the local adjacent areas in LR. Therefore, the dimension in $3$-th and $4$-th of the last self-matrix is $\frac{N}{s}\times \frac{N}{s}$, which can significantly reduce the computational and GPU memory cost. Denote these segments as $Q_1,...,Q_s$, $K_1,...,K_s$, and $V_1,...,V_s$. Each triplet of these segments is applied with a Scaled Dot-Product Attention (SDPA) operation, respectively. The structure of SDPA is also shown in Figure~\ref{fig:erst}, which just omits the $Mask$ operation. Afterward, all the outputs ($O_1, O_2, ..., O_s$) of SDPA are concatenated together to generate the whole output feature $O$. Finally, an Expansion layer is used to recover the number of channels.

\section{Experiments}
\subsection{Datasets and Metrics}
In this work, we use DIV2K~\cite{div2k} as the training dataset. For evaluation, we use five benchmark datasets, including Set5~\cite{Set5}, Set14~\cite{Set14}, BSD100~\cite{B100}, Urban100~\cite{huang2015single}, and Manga109~\cite{manga109}. Meanwhile, PSNR and SSIM are used to evaluate the performance of the reconstructed SR images.

\subsection{Implementation Details}
\textbf{Training Setting.} During training, we randomly crop $16$ LR image patches with the size of $48\times 48$ as inputs in each epoch. Random horizontal flipping and $90$ degree rotation is used for data augment. The initial learning rate is set to $2\times 10^{-4}$ and decreased half for every 200 epochs. The model is trained by Adam optimizer with a momentum equal to $0.9$. Meanwhile, $L1$ loss is used as it can produce more sharp images compared to $L2$ loss. Meanwhile, we implement ESRT with the PyTorch framework and train ESRT roughly takes two days with one GTX1080Ti GPU.

\textbf{Implements Details.} In ESRT, we set $3\times3$ as the size of all convolutional layers except for the Reduction module, whose kernel size is $1\times1$. Each convolutional layer has $32$ channels except for the fusion layer which is twice. For the image reconstruction part, following most previous methods, we use PixelShuffle~\cite{espcn} to upscale the last coarse features to fine features. The $k$ in HFP is set to $2$, which means that the feature map is down-scaled by half. Meanwhile, the number of HPB is set to $3$, the initial value of learnable weight in ARFB is set to $1$, and the number of ET in LTB is set to $1$ to save the GPU memory. In addition, the splitting factor $s$ in ET is set to $4$, the $k$ in pre- and post-process of ET is set to $3$, and the head number $m$ in EMHA is set to $8$, respectively.

\begin{figure*}
\centering
\includegraphics[scale=0.15]{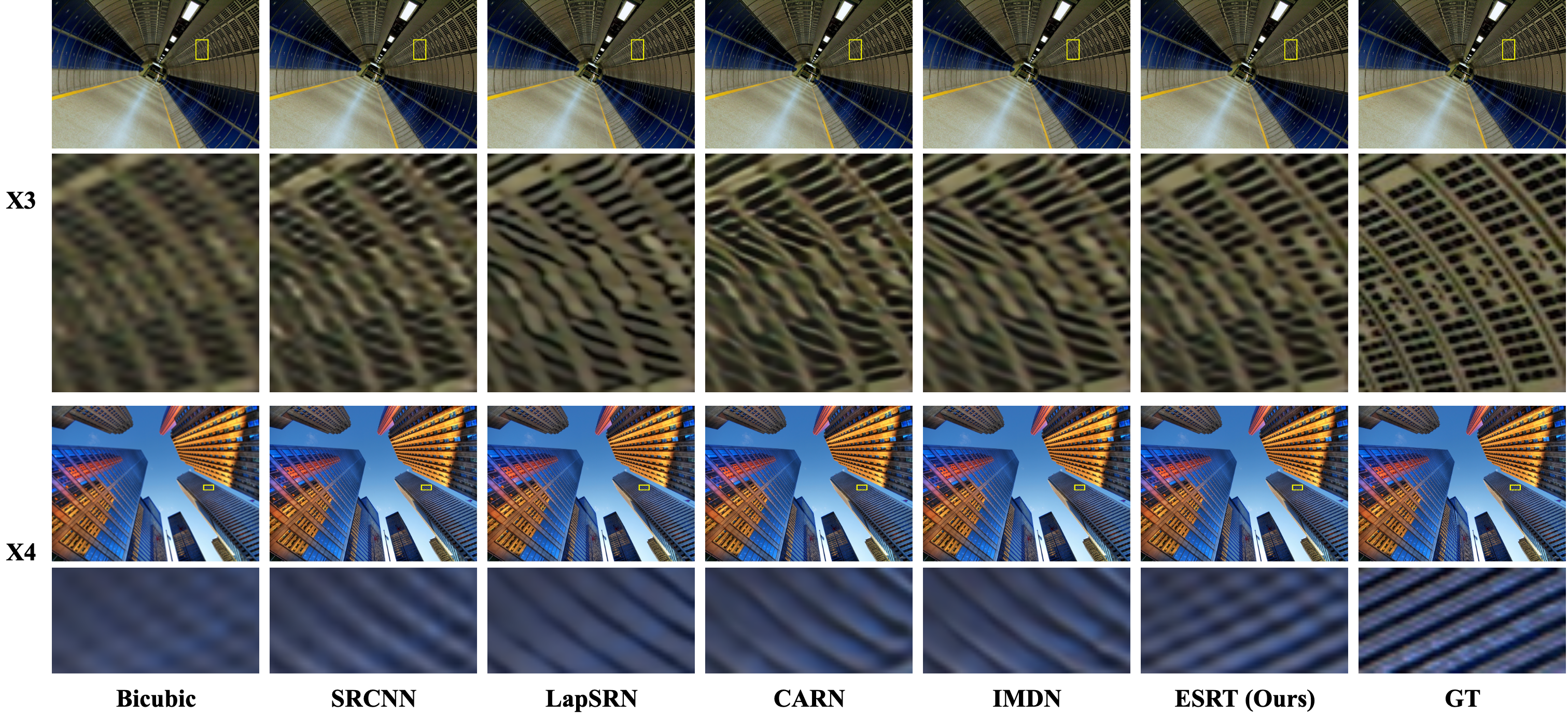}
\caption{Visual comparison with lightweight SISR models. Obviously, ESRT can reconstruct realistic SR images with sharper edges.}
\label{fig:vis-imgs}
\vspace{-0.3cm}
\end{figure*}

\subsection{Comparisons with Advanced SISR Models}
In TABLE~\ref{tab:psnr-ssim}, we compare ESRT with other advanced SISR models. Obviously, our ESRT achieves competitive results under all scaling factors. It can be seen that although the performance of EDSR-baseline is close to ESRT, it has almost twice as many parameters as ESRT. Meanwhile, the number of MAFFSRN and LatticeNet is close to ESRT, but ESRT achieves better results than them. Moreover, we can observe that our ESRT performs much better than other models on Urban100. This is because there are many similar patches in each image of this dataset. Therefore, the introduced LTB in our ESRT can used to capture the long-term dependencies among these similar image patches and learn their relevance, thus achieve better results.

In Figure~\ref{fig:vis-imgs}, we also provide the visual comparison between ESRT and other lightweight SISR models on $\times 2$, $\times 3$, and $\times 4$. Obviously, SR images reconstructed by our ESRT contains more accurate texture details, especially in the edges and lines. It is worth noting that in the $\times 4$ scale, the gap between ESRT and other SR models is more apparent. This benefits from the effectiveness of the proposed Efficient Transformer, which can learn more information from other clear areas. All these experiments validate the effectiveness of the proposed ESRT.

\subsection{\textbf{Comparison on Computational Cost}} 
\begin{figure}
  \begin{center}
  \includegraphics[scale=0.35]{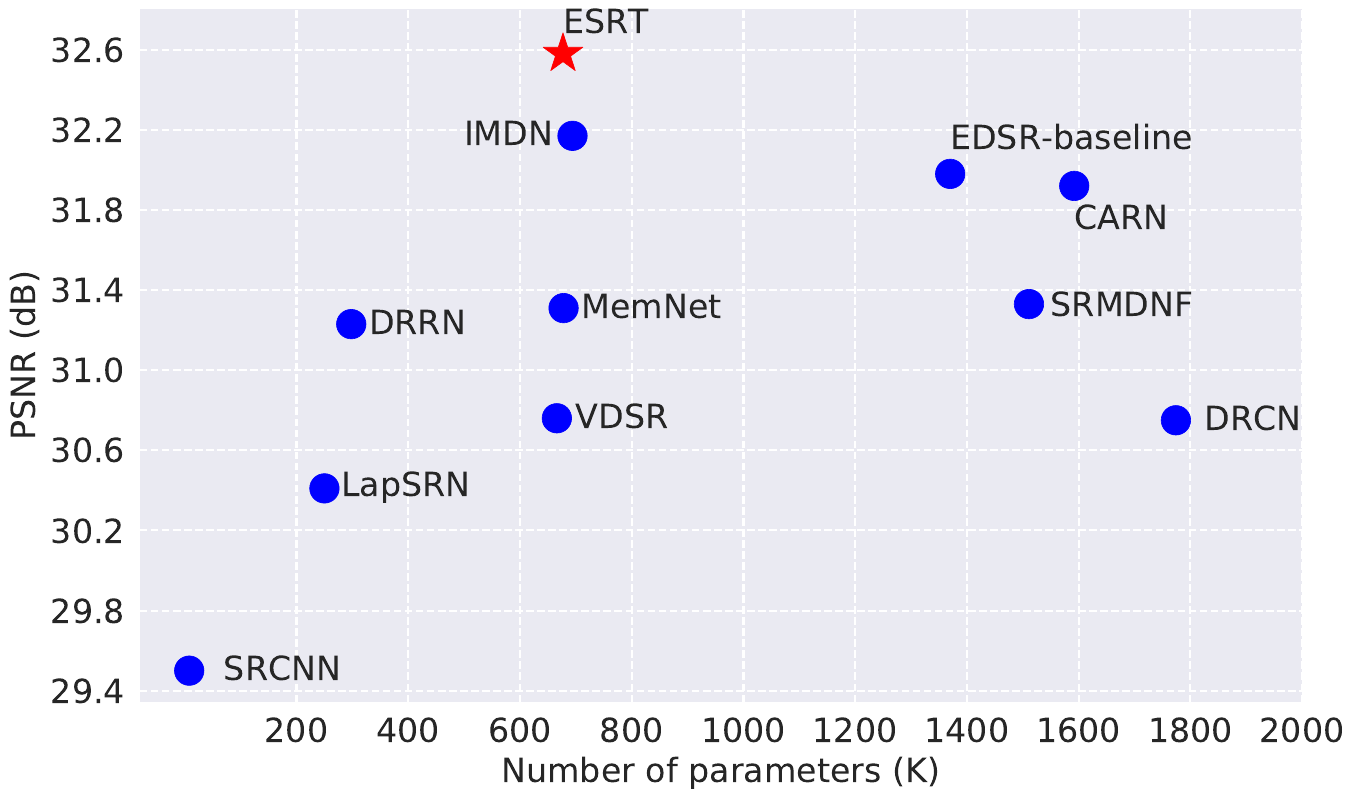}
  \end{center}
  \caption{Study the trade-off between the number of model parameters and performance on Urban100 ($\times 2$).}
  \label{fig:trade}
  \vspace{-0.3cm}
\end{figure}

In TABLE~\ref{tab:layers}, we provide a more detailed comparison of each model. It can be seen that ESRT can achieve 163 layers while still achieves the second-least FLOPs (67.7G) among these methods. This is benefited from the proposed HPB and ARFB, which can efficiently extract useful features and preserving the high-frequency information. Meanwhile, we can observe that the execution time is short even though ESRT uses the Transformer architecture. The increased time is completely acceptable compared to CARN and IMDN. In addition, we also visualize the trade-off analysis between the number of model parameters and performance in Figure~\ref{fig:trade}. Obviously, we can see that our ERST achieves a good trade-off between the size and performance of the model.

\begin{table}
  \centering
  \small
  \setlength{\tabcolsep}{0.8mm}
  \renewcommand{\arraystretch}{0.88}
  \begin{tabular}{l|ccccc}
    \toprule[1pt]
     Method &Layers &RL & Param. & FLOPs (x4) & Running time \\
     \hline
     VDSR~\cite{vdsr}               &20  &Yes & 0.67M& 612.6G & 0.00597s \\
     LapSRN~\cite{LapSRN}           &27  &Yes &0.25M& 149.4G & 0.00330s \\
     DRRN~\cite{drrn}               &52  &No  &0.30M& 6796.9G & 0.08387s\\
     
     CARN~\cite{CARN}               &34  &Yes &1.6M&  90.9G & 0.00278s\\
     IMDN~\cite{imdn}               &34  &Yes &0.7M& 40.9G & 0.00258s\\
     \hline
     \bf{ESRT}            & 163 &Yes &0.68M& 67.7G & 0.01085s \\
     \bottomrule[1pt]
  \end{tabular}
  \caption{Network structure settings comparison between our ESRT and other lightweight SISR models (the input size is $1280\times 720$).}
  \label{tab:layers}
  \vspace{-0.3cm}
 \end{table}

 \begin{table}
  \centering
  \small
	\setlength{\tabcolsep}{2mm}
  \renewcommand\arraystretch{0.88}
	\begin{tabular}{c|cccc}
    \toprule[1pt]
    Case Index &1&2&3&4\\
    \hline
    HFM     &         &$\surd$ &$\surd$ &$\surd$ \\
    CA      & $\surd$ &$\surd$ &        &$\surd$ \\
    ARFB    & $\surd$ &$\surd$ &$\surd$ &        \\
    RB      &         &        &        &$\surd$ \\
    \hline
    Parameters & 658K     & 751K    & 724K    &972K        \\
    PSNR      & 32.02dB   & \underline{32.19dB}  & 32.08dB  & \textbf{32.20dB}        \\
    \bottomrule[1pt]
  \end{tabular}
  \caption{Study of each component in HPB on Set5 ($\times 4$).}
  \label{tab:hpb}
  \vspace{-0.3cm}
\end{table}

\subsection{Network Investigations}
\subsubsection{\textbf{Study of High Preserving Block (HPB)}}
HPB is an important component of ESRT, which not only can reduce the model size but maintain the high performance of the model. To prove it, we explore the effectiveness of each component of ESRT in TABLE~\ref{tab:hpb}. According to Cases 1, 2, and 3, we can observe that the introduced HFM and CA can effectively improve the model performance at the cost of a few parameters. According to Cases 2 and 4, we can see that if RB is used to represent ARFB, the PSNR result just rises 0.01dB but the number of parameters go up to 972K. This means that ARFB can significantly reduce the model parameters while maintaining excellent performance. All these results fully illustrate the necessity and effectiveness of these modules and mechanisms in HPB.

\subsubsection{\textbf{Study of Efficient Transformer (ET)}}
To capture the long-term dependencies of similar local regions in the image, we introduced the Transformer and proposed a Efficient Transformer (ET). In TABLE~\ref{tab:et}, we analyze the model with and without Transformer. It can be see that if ESRT removes the Transformer, the model performance descends obviously from 32.18dB to 31.96dB. This is because the introduced Transformer can make full advantage of the relationship between similar image patches in an image. In addition, we compare our ET with the original Transformer~\cite{ViT} in the table. Our model (1ET) achieves better results with fewer parameters and GPU memory consumption (1/4). This experiment fully verified the effectiveness of the proposed ET. Meanwhile, we can also see that when the number of ET increases, the model performance will be further improved. However, it is worth noting that the model parameters and GPU memory will also increase when the number of ET increases. Therefore, to achieve a good balance between the size and performance of the model, only one ET is used in the final ESRT.

\begin{table}
  \centering
  \small
	\setlength{\tabcolsep}{1.5mm}
  \renewcommand\arraystretch{0.88}
	\begin{tabular}{c|ccc}
    \toprule[1pt]
    Case                   &PSNR(dB)  &Parame.(K)&GPU Memory\\
    \hline
    w/o TR                 & 31.96       &554        & 1931M          \\
    Original TR~\cite{vaswani2017attention}            & 32.14       &971        & 16057M \\
    1 ET                  &\textbf{32.18}        &751        & 4191M  \\
    2 ET                   & 32.25       &949        & 6499M       \\
    \bottomrule[1pt]
  \end{tabular}
  \caption{Study of Efficient Transformer (ET) on Set5 ($\times 4$). The GPU memory here refers to the cost of the model during training, which patch\_size = 48*48 and batch\_size=16.}
  \label{tab:et}
  \vspace{-0.3cm}
\end{table}

\begin{table}
  \small
  \centering
  \setlength{\tabcolsep}{1.2mm}
  \renewcommand{\arraystretch}{0.88}
	\begin{tabular}{c|c|c|cccc}
    \toprule[1pt]
    Scale& Model&Param           &Set5  &Set14   &Urban100\\
    \hline
    \multirow{2}{*}{$\times 3$}&RCAN~\cite{RCAN}&16M      &\textbf{34.74dB} &\textbf{30.65dB}  &29.09dB\\
                               &RCAN/2+ET&8.7M &34.69dB &30.63dB    &\textbf{29.16dB}\\
    \hline
    \multirow{2}{*}{$\times 4$}&RCAN~\cite{RCAN}&16M &\textbf{32.63dB} &28.87dB  &26.82dB\\
                               &RCAN/2+ET&8.7M &32.60dB &\textbf{28.90dB}  &\textbf{26.87dB}\\
  \bottomrule[1pt]
  \end{tabular}
  \caption{Comparison between RCAN and RCAN/2+ET.}
  \label{tab:rcan}
  \vspace{-0.3cm}
\end{table}

In order to verify the effectiveness and universality of the proposed ET, we also introduce ET into RCAN~\cite{RCAN}. It is worth noting that we use a small version of RCAN (the residual group number is set to $5$) and add the ET before the reconstruction part. According to TABLE~\ref{tab:rcan}, we can see that the performance of the model ``RCAN/2+ET" is close or even better than the original RCAN with fewer parameters. This further proves the effectiveness and universality of ET, which can be easily transplanted to any existing SISR models to further improve the performance of model.

\subsection{\textbf{Real Image Super-Resolution}}
To further verify the validity of the model, we also compare our ESRT with some classic lightweight SR models on the real image dataset (RealSR~\cite{realsr}). According to TABLE~\ref{tab:esrt-real}, we can observe that ESRT achieves better results than IMDN. In addition, ESRT achieves better performance than LK-KPN on $\times 4$, which was specifically designed for the real SR task. This experiment further verifies its effectiveness on real images.

\subsection{\textbf{Comparison with SwinIR}}
The EMHA in our ESRT is similar to the Swin Transformer layer of SwinIR~\cite{liang2021swinir}. However, SwinIR uses a sliding window to solve the high computation problem of the Transformer while ESRT uses a splitting factor to reduce the GPU memory consumption. According to Table~\ref{tab:Swin}, compared with SwinIR, ESRT achieves close performance with fewer parameters and GPU memory. It is worth noting that SwinIR uses an extra dataset (Flickr2K~\cite{agustsson2017ntire}) for training, which is the key to further improving the model performance. For a fair comparison with methods such as IMDN, we did not use this external dataset in this work.

\begin{table}
  \small
    \setlength{\tabcolsep}{2mm}
    \renewcommand\arraystretch{0.88}
    \centering
    \begin{tabular}{l | c c | c c | c c }
      \toprule[1pt]
        \multirow{2}{*}{Scale}& \multicolumn{2}{c|}{IMDN~\cite{imdn}}& \multicolumn{2}{c|}{LK-KPN~\cite{realsr}}& \multicolumn{2}{c}{ESRT (Ours)}\\
         & PSNR & SSIM & PSNR & SSIM & PSNR & SSIM\\
        \midrule
        $\times$3 & 30.29  & 0.857 & \textbf{30.60}  & \textbf{0.863} &30.38 & 0.857\\
        $\times$4 & 28.68  &  0.815 & 28.65 & \textbf{0.820} &\textbf{28.78}&0.815\\
      \bottomrule[1pt]
      \end{tabular}
    \vspace{-0.2cm}
    \caption{Comparison with advanced SISR methods on RealSR.}
    \label{tab:esrt-real}
    \vspace{-0.25cm}
\end{table}

\begin{table}
  \centering
  \small
	\setlength{\tabcolsep}{1.2mm}
  \renewcommand\arraystretch{0.88}
	\begin{tabular}{c|cccc}
    \toprule[1pt]
    Method  &Parame.  &GPU Memory &BSD100 &Manga109\\
    \hline
    SwinIR  & 886K    &6966M    & 29.20/0.8082 & 33.98/0.9478 \\
    ESRT    & 770K    &4191M    & 29.15/0.8063 & 33.95/0.9455 \\
    \bottomrule[1pt]
  \end{tabular}
  \vspace{-0.2cm}
  \caption{A detailed comparison of SwinIR and ESRT ($\times 4$).}
  \label{tab:Swin}
  \vspace{-0.5cm}
\end{table}

\section{Conclusion}
In this work, we proposed a novel Efficient Super-Resolution Transformer (ESRT) for SISR. ESRT first utilizes a Lightweight CNN Backbone (LCB) to extract deep features and then uses a Lightweight Transformer Backbone (LTB) to model the long-term dependence between similar local regions in an image. In LCB, we proposed a High Preserving Block (HPB) to reduce the computational cost and retain high-frequency information with the help of the specially designed High-frequency Filtering Module (HFM) and Adaptive Residual Feature Block (ARFB). In LTB, an Efficient Transformer (ET) is designed to enhance the feature representation ability with lower GPU memory occupation under the help of the proposed Efficient Multi-head Attention (EMHA). Extensive experiments demonstrate that ESRT achieves the best trade-off between model performance and computation cost.

\section{Acknowledgment}
This work was supported by National Key R\&D Program of China (2021YFE0203700), CRF (8730063), National Natural Science
Foundation of China (U1613209), and Science and Technology Plan Projects of Shenzhen (JCYJ20190808182209321, JCYJ20200109140410340).

{\small
\bibliographystyle{ieee_fullname}
\bibliography{egbib}
}

\end{document}